\documentclass{article}

\usepackage[final, nonatbib]{tackling_climate_workshop_style}
\usepackage[utf8]{inputenc} % allow utf-8 input
\usepackage[T1]{fontenc}    % use 8-bit T1 fonts
\usepackage{hyperref}       % hyperlinks
\usepackage{url}            % simple URL typesetting
\usepackage{amsfonts}       % blackboard math symbols
\usepackage{nicefrac}       % compact symbols for 1/2, etc.
\usepackage{microtype}      % microtypography
\usepackage{graphicx}
\usepackage{multirow}
\usepackage{appendix}
\usepackage{latexsym}
\usepackage{mdframed}
\usepackage{floatrow}
\newfloatcommand{capbtabbox}{table}[][\FBwidth]
\usepackage{blindtext}

\title{\textsc{ClimateX}: Do LLMs Accurately Assess Human Expert Confidence in Climate Statements?}

\author{%
 Romain Lacombe \\
 Stanford University \\
 \texttt{rlacombe@stanford.edu} \\\And
 Kerrie Wu \\
 Stanford University \\  
 \texttt{kerriewu@stanford.edu} \\\And
 Eddie Dilworth \\
 Stanford University \\  
 \texttt{edjd@stanford.edu}
 }

\begin{document}

\maketitle

\begin{abstract}

Evaluating the accuracy of outputs generated by Large Language Models (LLMs) is especially important in the climate science and policy domain. We introduce the Expert Confidence in Climate Statements (\textsc{ClimateX}) dataset, a novel, curated, expert-labeled dataset consisting of 8094 climate statements collected from the latest Intergovernmental Panel on Climate Change (IPCC) reports, labeled with their associated confidence levels. Using this dataset, we show that recent LLMs can classify human expert confidence in climate-related statements, especially in a few-shot learning setting, but with limited (up to 47\%) accuracy. Overall, models exhibit consistent and significant over-confidence on low and medium confidence statements. We highlight implications of our results for climate communication, LLMs evaluation strategies, and the use of LLMs in information retrieval systems.

\end{abstract}

%%%%%%%%%%%%%%%%%%%%%%%%%%%%%%%%%%%%%%%%%%%%

\section{Introduction}

The wide deployment of Large Language Models (LLMs) as question-answering tools calls for nuanced evaluation of their outputs across knowledge domains. This is especially important in climate science, where the quality of the information sources shaping public opinion, and ultimately public policy, could determine whether the world succeeds or fails in tackling climate change.

This paper aims to evaluate the reliability of LLM outputs in the climate science and policy domain. We introduce the \textbf{Expert Confidence in Climate Statements (\textsc{ClimateX}) dataset} \cite{climatex}, a novel, curated, expert-labeled, natural language dataset of 8094 statements sourced from the three most recent Intergovernmental Panel on Climate Change Assessment Reports (IPCC AR6) — along with their associated confidence levels (low, medium, high, or very high) that were assessed by climate scientists based on the quantity and quality of available evidence and agreement among their peers. \textsc{ClimateX} is available on \href{https://huggingface.co/datasets/rlacombe/ClimateX}{HuggingFace} and source code for experiments is available on \href{https://github.com/rlacombe/ClimateX}{Github}.

We use this dataset to evaluate how accurately recent LLMs assess the confidence which human experts associate with climate science statements. Although OpenAI's \texttt{GPT-3.5-turbo} and \texttt{GPT-4} assess the true confidence level with better-than-random accuracy and higher performance than non-expert humans, even in a zero-shot setting, they, and other models we tested, consistently overstate the certainty level associated with low and medium confidence labeled statements.

With LLMs poised to become increasingly significant sources of public information, the reliability of their outputs in the climate domain is critical for avoiding misinformation and garnering support for effective climate policy. We hope the \textsc{ClimateX} dataset provides a valuable tool for benchmarking the trustworthiness of LLM outputs in the climate domain, highlights the need for further work in this area, and aids efforts to develop models that accurately convey climate knowledge.

\section{Prior Work}

%%%%%%%%%%%%%%%%%%%%%%%%%%%%%%%%%%%%%%%%%%%%

\subsection{LLMs and Uncertainty}

Recent literature demonstrates that linguistic statements of certainty impact LLM performance on calibrated NLP tasks. For example, Zhou et al. (2023) \cite{Zhou2023} find that appending `weakeners' (i.e. ``A: I think it's...'') or `strengtheners' (i.e. ``A: I'm certain it's...'') to zero-shot QA prompts can significantly impact GPT-3's performance on common QA datasets like TriviaQA. They find that strengtheners surprisingly result in a lower accuracy across their datasets (40\%) than weakeners (47\%), and suggest that there may be unique challenges with reliably interpreting linguistic cues of high confidence.

To address this issue, and the more general over-confidence problem, recent work attempts to train LLMs to accurately express their own uncertainty. Lin et al. (2022) \cite{lin2022teaching} fine-tune \cite{howard2018universal} GPT-3 to express how confidently it answers different arithmetic tasks, using both categorical certainty (e.g. ``high confidence'') and numeric certainty (e.g. ``90\%''). While they obtain relative success and promising results, the authors note concerns such as over-fitting to their training set distribution.

Kadavath et al. (2022) \cite{kadavath2022language} show that LLMs are capable of `self-evaluation,' i.e. evaluating their own answers as true or false relative to accepted human knowledge with few-shot learning. They also fine-tune models to predict the probability that they can accurately answer a particular question. The authors train and evaluate many different-sized models using multiple pre-existing generation datasets such as TriviaQA, arithmetic, and code generation. They find that models are initially poor at self-evaluation, but improve with few-shot learning (up to 20 demonstrations).

Prior work has also explored how to evaluate ``what LLMs know.'' Chang et al. (2023) \cite{chang2023speak} carry what they describe as a ``data archaeology'' investigation to infer books that LLMs have been trained on, using a ``name cloze membership inference query,'' the task of predicting a masked name in a sentence based on the context surrounding it. Importantly, the human baseline on this task is 0\%. The authors find a clear correlation between the frequency at which books appear in datasets over which LLMs are known to have been trained, and their performance on the related cloze task. 

Finally, prior work has also included more wholistic evaluations of how LLMs communicate  climate-change related information. Bulian et al. (2023) \cite{bulian2023assessing} present a framework for evaluating how well LLMs convey climate-change related information. They prompt LLMs to provide a long-form (3-4 paragraphs) response to a climate-related question, and ask humans to rate LLM-generated completions along multiple axes such as accuracy, specificity, completeness, and most relevantly, communicating levels of uncertainty properly. 

These prior works suggest that tasking LLMs to assess confidence in climate statements in a zero-shot setting may be a challenging task, particularly for reports released after model knowledge cutoff dates, but it could be made more tractable by few-shot prompting with demonstrations.

\subsection{Uncertainty Language in IPCC Special Reports}

%%%%

The IPCC reports have proven to be an excellent context source for retrieval-augmented LLM systems, such as in chatClimate \cite{vaghefi2023chatclimate}, which showed the benefit of using the IPCC reports to ground conversational AI agents. Prior work also used climate science literature as a benchmark for NLP systems, such as \textsc{ClimaBench} \cite{laud2023climabench} and \textsc{ClimaText} \cite{varini2021climatext}, which evaluate LLMs' performance on a range of tasks like climate topic classifications and knowledge-related QA.

In 2010, the IPCC issued a set of guidelines \cite{Mastrandrea2010} to lead authors of the IPCC Reports on how to consistently communicate uncertainty. Janzwood \cite{Janzwood2020} analyzes the reports written after the guidelines were published and finds evidence of broad adoption across chapter authors and IPCC reports of the \textbf{Confidence} framework, which evaluates scientific confidence in each statement by the quantity and quality of available evidence and agreement among peers. Confidence is measured on a 5-level categorical scale including `very low', `low', `medium', `high', and `very high confidence'. 

This encouraged us to create the \textsc{ClimateX} dataset, comprising of sentences extracted from the IPCC AR6 reports and labeled according to the prevalent 5-levels confidence scale, to evaluate how accurately LLMs assess the confidence level attributed to climate statements by a consensus of human experts. \textsc{ClimateX} adds to a growing number of climate benchmarks and is, to our knowledge, \textbf{the first LLM climate benchmark to deliberately probe uncertainty, and how accurately these models assess human expert confidence in climate statements.}

%%%%%%%%%%%%%%%%%%%%%%%%%%%%%%%%%%%%%%%%%%%%

\section{\textsc{ClimateX}: the Expert Confidence in Climate Statements dataset}

We introduce \textbf{\textsc{ClimateX}, the Expert Confidence in Climate Statements dataset} \cite{climatex}, a novel, curated, expert-labeled, natural language dataset of 8094 statements sourced from the three latest IPCC reports (Assessment Report 6: Working Group I \cite{ipcc_wg1}, Working Group II \cite{ipcc_wg2}, and Working Group III \cite{ipcc_wg3}, respectively). Each statement is labeled with its source IPCC report and page number, and the corresponding confidence level on the 5-level confidence scale as assessed by IPCC climate scientists based on available evidence and agreement among their peers. 

To construct the dataset, we retrieved the complete raw text from each of the three IPCC report PDFs that are available online using an open-source library \cite{pypdf2}, normalized the whitespace, tokenized the text into sentences using NLTK \cite{bird2009natural}, and used regular expression search to filter for complete sentences including a parenthetical confidence label at the end of the statement, of the form:
$$\texttt{``<statement> (\{low|medium|high|very high\} confidence)''}$$ 

The complete \textsc{ClimateX} dataset contains \textbf{8094 labeled statements}. From these sentences, we randomly selected \textbf{300 statements} to form a \textbf{test dataset}, while the remainder form the \textbf{train split}. Test samples selection aimed to ensure the test set is: (i) representative of the full data set with regard to confidence class per source report distributions; (ii) representative of the train set with regard to the number of statements from each report; and (iii) contains at least five sentences from each class and each report. Then, we manually reviewed each sentence in the test set, and, where necessary, cleaned up or clarified terms based on paragraph context, to provide for a fairer assessment of model capacity.

We report the percentage of statements per confidence class and source report for the full data set in Figure \ref{fig:climatex_distribution}, and for the test set in Table \ref{tab:test-set}. Note that the data set excludes the `very low confidence' class because `very low confidence' statements are almost entirely absent from the final reports.

%%%%%%%%%%%%%%%%%%%%%%%%%%%%%%%%%%%%%%%%

\section{Using \textsc{ClimateX} to Evaluate LLMs}
\subsection{Prompting models for confidence in climate science statements}

We compared the performance of three recent commercial LLMs: OpenAI's \texttt{GPT-3.5-turbo}, \texttt{GPT-4}, and Cohere's \texttt{Command-XL} \cite{OpenAIModels2023, coherecommand}. In experiments completed between June 1 and June 9, 2023, we prompted publicly available APIs for each of these models with sentences from the test split of the \textsc{ClimateX} dataset, and instructed the model to attempt to predict the corresponding confidence labels using the template shown in Figure \ref{figure:zeroshotprompt} in Appendix \ref{sec:prompts}. 

Using the Demonstrate-Search-Predict (DSPy) library \cite{DSP}, we prompted models in both a \textbf{zero-shot setting} with no additional context, and a \textbf{few-shot learning setting} with four ground-truth demonstrations randomly selected from the train split of the \textsc{ClimateX} dataset  (one from each confidence class). Models were prompted to output one of the four confidence levels, or to communicate if they were unable to make a confidence classification due to knowledge limitations.

\subsection{Metrics: quantifying confidence levels}

\begin{figure}[!ht]
\centering
\includegraphics[width=0.6\linewidth]{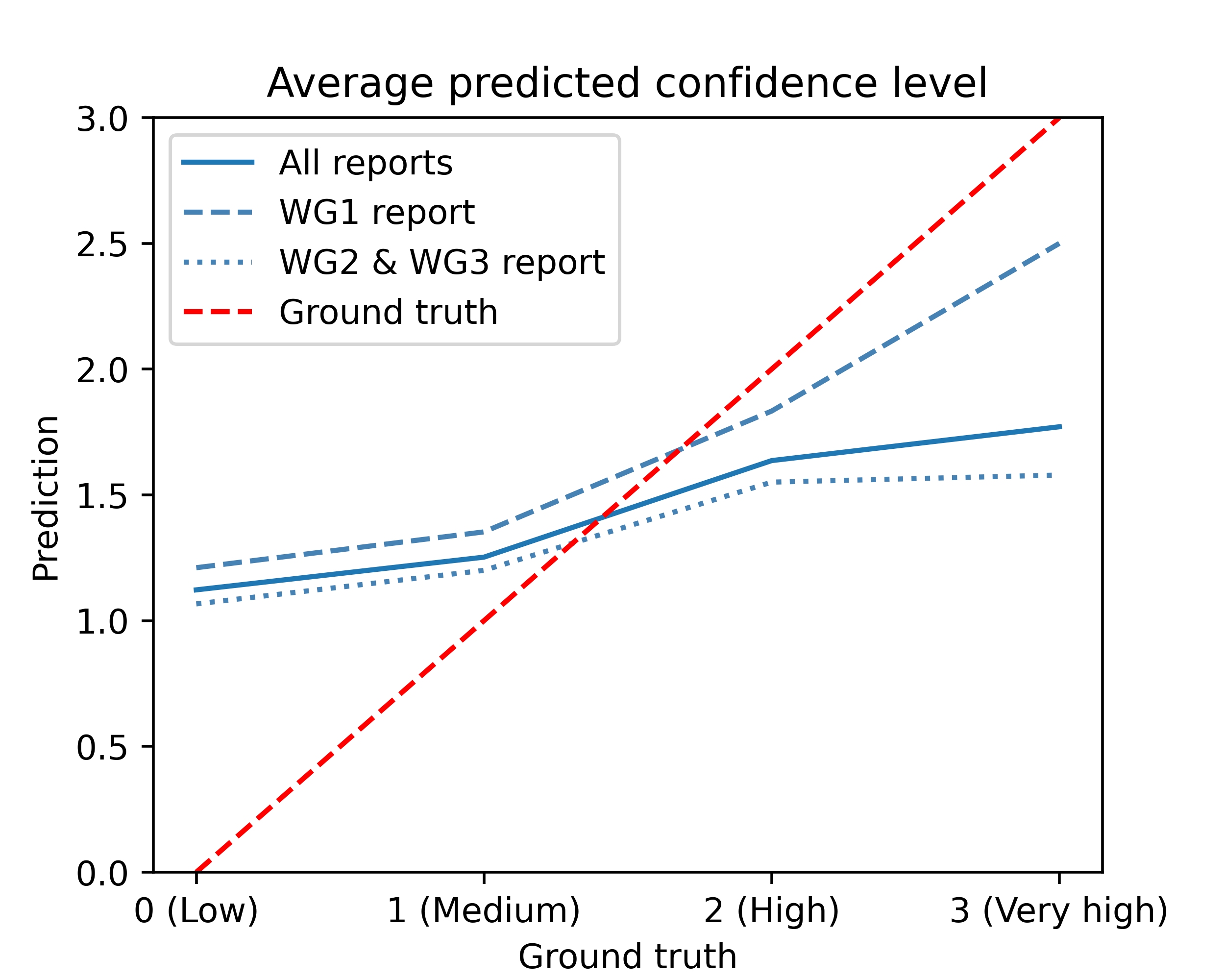}  
\caption{Average predicted confidence level per class. Setting: \texttt{GPT3.5-turbo}, few-shot.}
\label{fig:gpt3.5_fewshot}
\end{figure}

To facilitate analysis, we map categorical confidence levels to a numerical score, where `low' corresponds to 0, `medium' to 1, `high' to 2, and `very high' to 3. The (rare) responses that do not fit into the four confidence level options are excluded from calculations. This mapping allows us to treat our LLM classifiers as regression models, and assess their performance by contrasting predicted confidence levels with ground truth as in a regression setting. We report our results as follows:
\begin{itemize}
    \setlength\itemsep{0.1em}
    \item Table \ref{tab:results} compiles the slope and bias of our regression of classifier outputs, for a view of the discernment (how well it distinguishes classes) and over/under-confidence of each model;
    \item Figure \ref{fig:gpt3.5_fewshot} plots the average confidence level predicted by \texttt{GPT-3.5-turbo} in the few shot setting for statements from each ground truth class (solid line), breaks down this result for statements from pre-knowledge cutoff date WGI report (dashed line) vs. post-cutoff WGII/III reports (dotted line), and compares it with a perfect classifier (red dashed line);
    \item Appendix \ref{sec:confidence-results} reports plots for each experiment in Figure \ref{fig:all-models-plots}, and underlying data in Table \ref{tab:confidence-results};
    \item Table \ref{tab:classifier-results} in Appendix \ref{sec:classifier-results} compiles classifier performance results for each experiment.
\end{itemize}

%%%%%%%%%%%%%%%%%%%%%%%%%%%%%%%%%%%%%%%%

%%%%%%%%%%%%%%%%%%%%%%%%%%%%%%%%%%%%%%%%

\begin{table}[!ht]
\centering
\resizebox{0.8\textwidth}{!}{%
\begin{tabular}{cccccccc}

\multicolumn{2}{c}{\textbf{Models}} & \multicolumn{2}{c}{\textbf{GPT-3.5-turbo}} & \multicolumn{2}{c}{\textbf{GPT-4}} & \multicolumn{2}{c}{\textbf{Cohere Command XL}} \\

\multicolumn{2}{c}{Setting} & Zero-shot & Few-shot & Zero-shot & Few-shot & Zero-shot & Few-shot \\ \hline \hline

\multirow{2}{*}{WGI} & Slope & 0.368 & \underline{\textbf{0.435}} & 0.399 & \underline{\textbf{0.445}} & 0.139 & \underline{\textbf{0.301}}  \\
 & Bias & +0.198 & +0.224 & +0.354 & +0.255 & +0.826 & +0.746  \\ \hline
\multirow{2}{*}{WGII/III} & Slope& 0.184 & 0.189 & 0.200 & 0.297 & 0.060 & 0.239  \\
 & Bias & -0.136 & -0.151 & -0.247 & +0.023 & +0.664 & +0.661  \\ \hline
\multirow{2}{*}{Total} & Slope& 0.215 & 0.233 & 0.223 & 0.323 & 0.069 & 0.257  \\
 & Bias & -0.046 & -0.054 & +0.042 & +0.083 & +0.708 & +0.683  \\ \hline
\end{tabular}%
}
\caption{Regression of average score per category against ground truth.
\\ \\
     \textit{\textbf{How to read this table?} (i) Slope: how much the average predicted score varies between statements of adjacent ground truth confidence classes. It measures how discerning the model is. Perfect classifier: 1. Random classifier: 0. (ii) Bias: difference between average predicted confidence and ground truth. It measures any bias towards over-confidence or over-caution. Unbiased classifier: 0. Over-confident by one class (classifies `medium' as `high'): +1: Over-cautious by one class (classifies `medium' as `low'): --1. 
}}
\label{tab:results}
\end{table}

%%%%%%%%%%%%%%%%%%%%%%%%%%%%%%%%%%%%%%%%

\section{Results and Analysis}

We present a summary of our results in Table \ref{tab:results}, and an illustration of the confidence classification performance of \texttt{GPT3.5-turbo} with few shots demonstrations in Figure \ref{fig:gpt3.5_fewshot}. We report detailed results for all models in Figure \ref{fig:all-models-plots} and Table \ref{tab:classifier-results} (Appendix \ref{sec:classifier-results}), which lead to the following observations.

\textbf{(1) Recent LLMs can classify human expert confidence in climate-related statements, but do so with limited accuracy.} \texttt{GPT-4} can classify statements with up to 44.3\% accuracy in the zero-shot learning setting, and 47.0\% in the few-shot setting (Table \ref{tab:classifier-results}). This is slightly better than the 36.3\% achieved by a small sample of 3 non-expert humans (Appendix \ref{sec:robustness_check}). All models exhibit a positive correlation between ground truth and predicted confidence level, with a slope varying from 0.06 for \texttt{Command-XL} in a zero-shot setup on WGII/WGIII reports (near random classifier), up to 0.445 for \texttt{GPT-4} in the few-shot in-context learning setting on the WGI report. 

\textbf{(2) Some models are consistently overconfident.} Cohere's \texttt{Command-XL}, in contrast to \texttt{GPT-3.5-turbo} and \texttt{GPT-4}, does particularly poorly by making strongly overconfident classifications (+0.708 bias). Even though \texttt{GPT-3.5-turbo} and \texttt{GPT-4} perform better overall (biases less than +0.1), they are not perfect---\textbf{all three models consistently over-estimate confidence in the `low' and `medium' categories within the \textsc{ClimateX} dataset (mean score > 1.0)}.

\textbf{(3) Few-shot learning significantly improves performance for all models.} When few-shot demonstrations are sampled from the train set to add context to the prompt, the slopes reported in Table \ref{tab:results} improve for all models, away from 0 (random classifier) and towards 1 (perfect classifier).

\textbf{(4) Models very rarely convey knowledge limitations}. `Support' in Table \ref{tab:confidence-results} reports the number of valid confidence classification, and shows that models convey knowledge limitation (``I don't know'') for none to at most 4\% of prompts. However, we caution that our prompt  begins with, "You are a knowledgeable climate science assistant trained to assess the confidence level associated with various statements," which could influence the models' knowledge-limitation outputs. We conducted robustness checks by prompting non-expert humans with the same task on the  \textsc{ClimateX} test set, and by prompting LLMs on nonsensical sentences (reported in Appendix \ref{sec:robustness_check}). 

\textbf{(5) Models are significantly better at discerning confidence levels for statements in the WGI report (released before the GPT-3.5-turbo knowledge cutoff date) than in the WGII/WGIII reports (released afterwards).}  As reported in Table \ref{tab:results}, the slope of average prediction scores relative to ground truth is about twice as high on statements from WGI than from WGII/III. Interestingly, this effect is also slightly visible in Cohere AI's \texttt{Command-XL}. 

Because the training sets for the models we tested are not disclosed, we cannot determine whether this is an emerging capability of LLMs or simply accurate recall of pre-training data. The discrepancy in model performance on the WGI report before the \texttt{GPT-3.5-turbo} knowledge cutoff date, compared to the WGII and WGIII reports that were published afterwards, strongly suggests the latter. We cannot exclude either possibility without more transparency on training procedures and datasets.

We also caution that each report covers different topics: the physical scientific basis for climate change (WGI), socio-economic and natural systems vulnerability to climate change (WGII), and climate mitigation options (WGIII). Some performance differences could be explained by relative exposure to literature on each topic during training, rather than recall of the specific IPCC language. 

%%%%%%%%%%%%%%%%%%%%%%%%%%%%%%%%%%%%%%%%

\section{Conclusion and Future Work}

Our findings add to the body of work calling to accelerate research on improving LLMs' communication of knowledge limitations to users, and calibration of their confidence assessment of statements (from themselves or others). Avoiding `hallucinations' from `confidently wrong' LLMs will be key for language model-based applications to function effectively as knowledge retrieval systems. An ability to generate outputs that convey confidence and certainty levels adequately is paramount for LLMs to meet that objective.

This is especially true regarding climate science and the communication of climate knowledge, which shapes public policy and impacts support for climate solutions. In the context of rising political polarization in public policy debates on climate change, where accurate and nuanced climate science communication is crucial, widely-deployed LLMs with the shortcomings highlighted here could exacerbate the spread of misinformation.

Areas for future work to improve and better understand LLM performance on classifying confidence levels associated with climate science statements include: (i) exploring better retrieval methods to select examples or context to include in the prompt; (ii) assessing open-source LLMs on the \textsc{ClimateX} test set, before and after fine-tuning them on the train set; (iii) probing whether natural language cues and qualifiers impact model performance on this task; and (iv) conducting a baseline human expert performance assessment on the \textsc{ClimateX} dataset.

%%%%%%%%%%%%%%%%%%%%%%%%%%%%%%%%%%%%%%%%

\section*{Acknowledgements}
The authors wish to thank Prof. Christopher Potts and Dr. Mina Lee at Stanford University for guidance on this project. We are also grateful to the thousands of climate scientists and experts who collaborated on the IPCC AR6 reports, and the many more whose research informed them. This work and the dataset we constructed would not have been possible without them.

%%%%%%%%%%%%%%%%%%%%%%%%%%%%%%%%%%%%%%%%

\section*{Dataset and code}
The \textsc{ClimateX} dataset is available for download on \href{https://huggingface.co/datasets/rlacombe/ClimateX}{HuggingFace}. 

Code for experiments reproduction and data analysis is available on \href{https://github.com/rlacombe/ClimateX}{Github}.

%%%%%%%%%%%%%%%%%%%%%%%%%%%%%%%%%%%%%%%%

%\clearpage
\bibliographystyle{plain}
\bibliography{anthology}

%%%%%%%%%%%%%%%%%%%%%%%%%%%%%%%%%%%%%%%%

\appendix 

\clearpage
\onecolumn

\section{Appendix: IPCC Guidelines to Authors on Confidence and Uncertainty Communication}\label{sec:ipcc}

In this section we present the IPCC guidelines to authors, as illustrated in the Working Group I Assessment Report 6 \cite{ipcc_wg1}.

\begin{figure*}[!ht]
\begin{center}
    
\includegraphics[width=\textwidth]{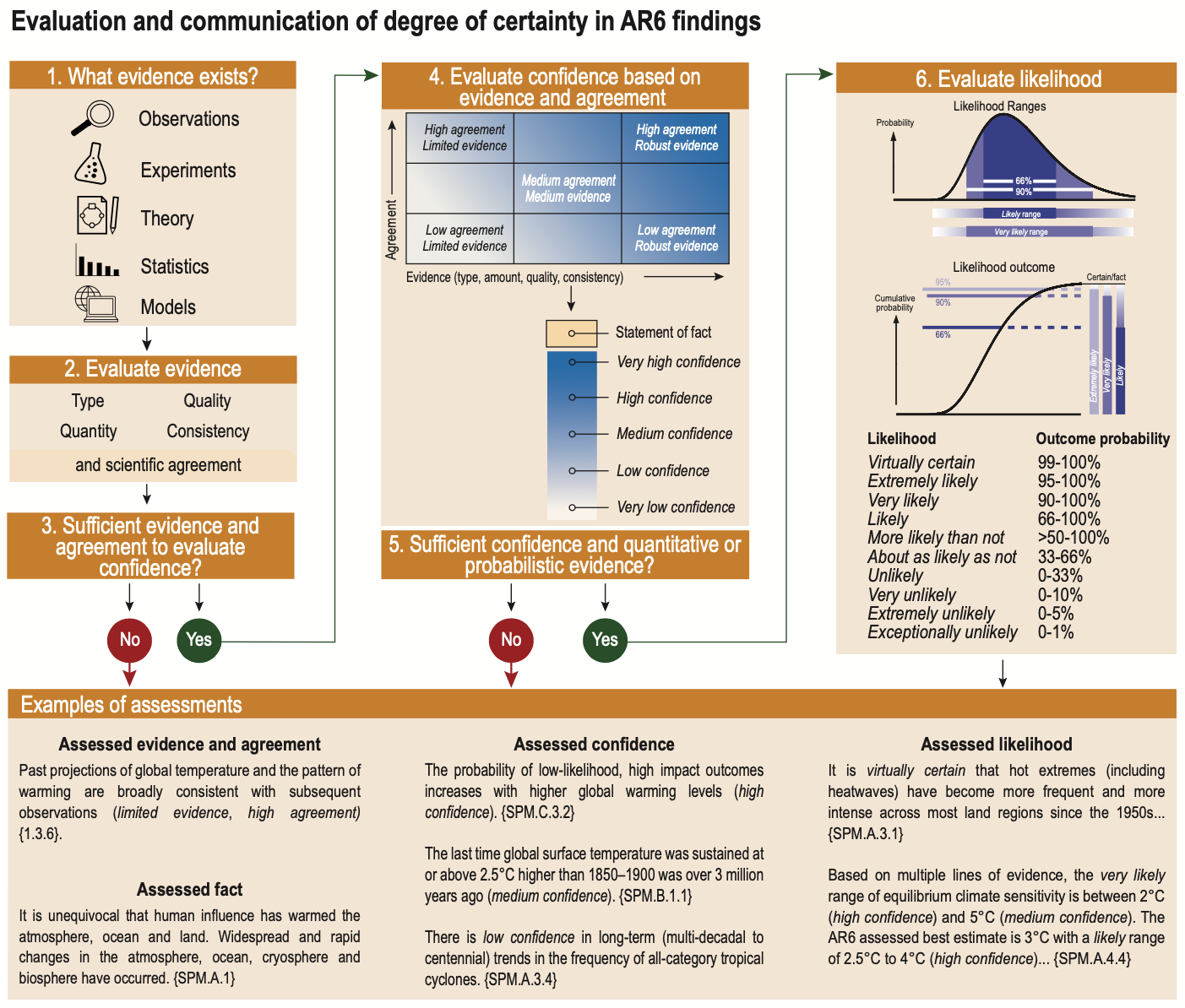}
\caption{IPCC guidelines to authors on communicating uncertainty and confidence.}
\end{center}

\label{fig:ipcc-scale}
\end{figure*}

\clearpage
\onecolumn

\section{Appendix: \textsc{ClimateX} Dataset Confidence Distribution}

\begin{figure}[hb]
\capbtabbox{%
\resizebox{0.6\textwidth}{!}{%
\begin{tabular}{lccccc}
Report \hspace{0.5em} & Low  & Medium  & High  & Very high  & Total \\ 
\hline \hline
WGI & 20 & 35 & 30 & 10 & 95 \\ \hline
WGII & 25 & 55 & 55 & 35 & 170 \\ \hline
WGIII & 5 & 10 & 15 & 5 & 35 \\ \hline
Total & 50 & 100 & 100 & 50 & 300
\end{tabular}%
}}{%
\caption{Test set: breakdown by confidence level and source IPCC report.}
\label{tab:test-set}
}
\end{figure}

\begin{figure}[htp]
\includegraphics[width=0.7\linewidth]{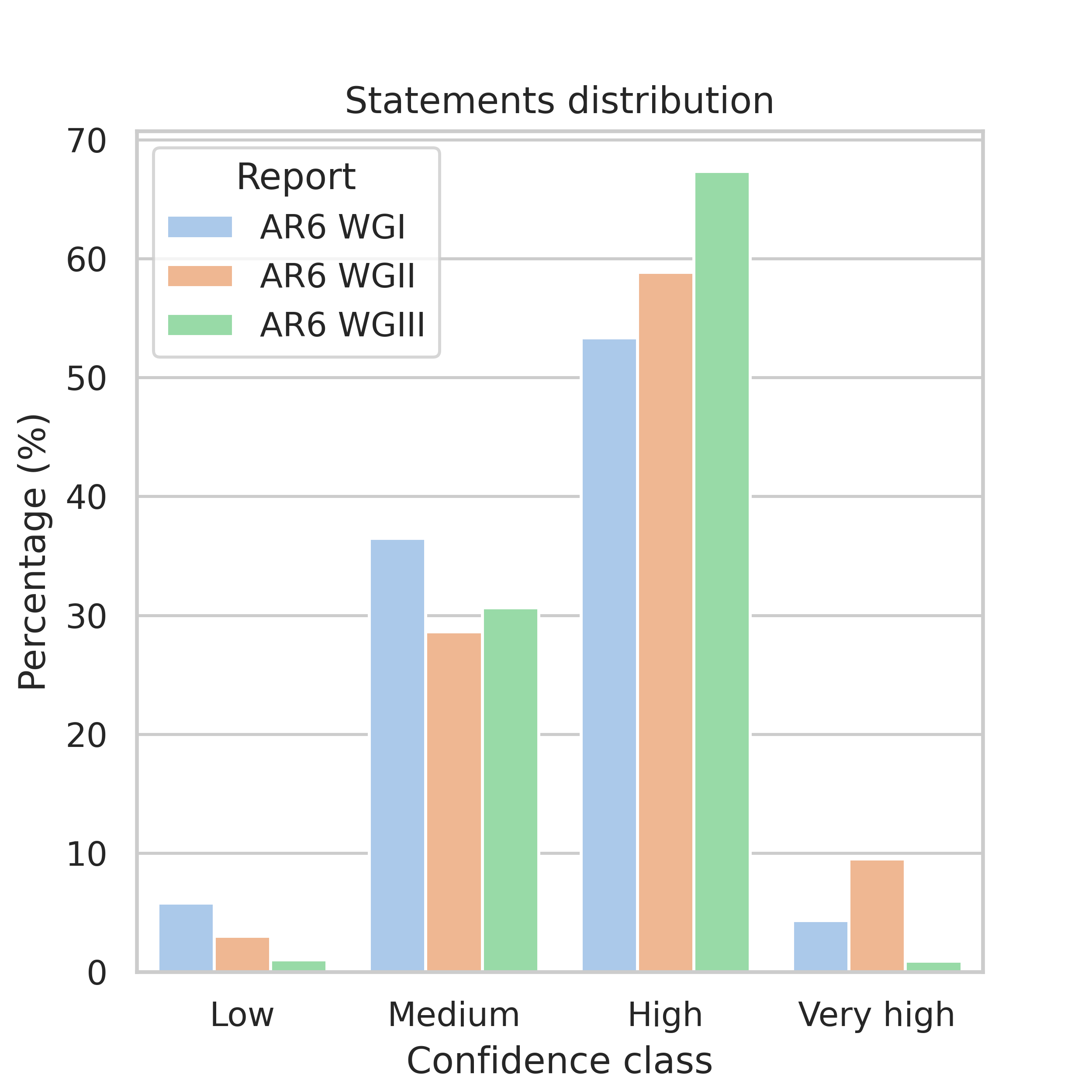}  

\caption{Percentage of statements corresponding to each confidence level and IPCC report source in the \textsc{ClimateX} dataset.}
\label{fig:climatex_distribution}
\end{figure}

\clearpage
\section{Appendix: Confidence Results Tables and Figures}\label{sec:confidence-results}

In this section we present the detailed confidence label predictions results for all experiments.

\begin{figure}[!ht]
\begin{center}
    
\includegraphics[width=\textwidth]{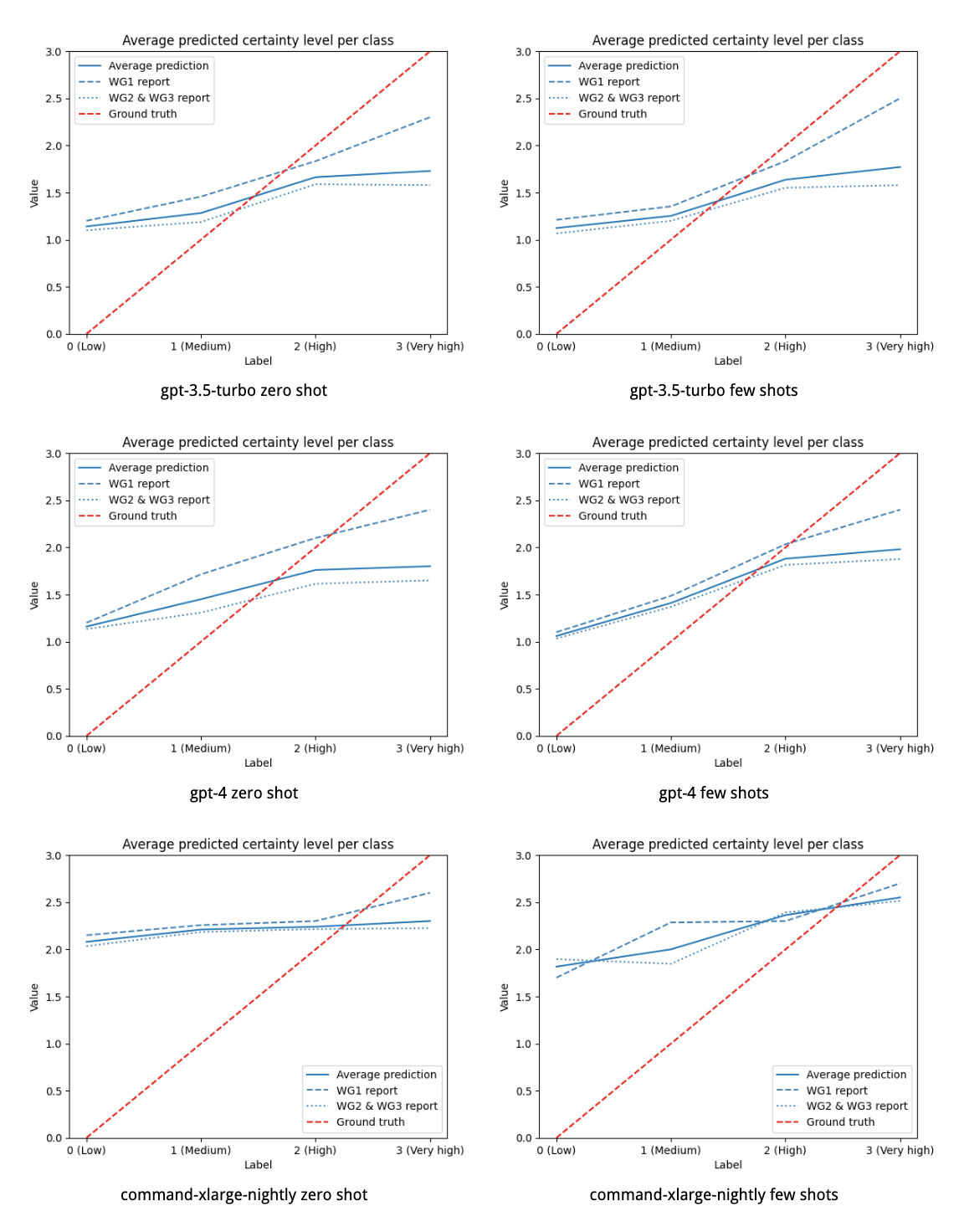}
\caption{Plots of average confidence level predictions per ground truth class label for all models in all settings.}
\label{fig:all-models-plots}
\end{center}

\end{figure}

\begin{table}[!ht]
\centering
\resizebox{1\textwidth}{!}{%
\begin{tabular}{llcccccc}

\multicolumn{2}{l}{\textbf{Models}} & \multicolumn{2}{c}{\textbf{GPT-3.5-turbo}} & \multicolumn{2}{c}{\textbf{GPT-4}} & \multicolumn{2}{c}{\textbf{Cohere Command XL}}\\

\multicolumn{2}{l}{Setting} & Zero-shot & Few-shot & Zero-shot & Few-shot & Zero-shot & Few-shot \\ \hline \hline

\multirow{4}{*}{`very high' (3)} & WG1        & 2.30     & 2.50           & 2.40            & 2.40               & 2.60                            & 2.70  \\
 & WG2\&3                                & 1.59       & 1.58           & 1.65          & 1.88              & 2.23                         & 2.51  \\
 & All reports                            & 1.73         & 1.77            & 1.80          & 1.98               & 2.30                        & 2.55 \\ 
 & Support                                & 48        & 48           & 50                & 50               & 50                          & 49 \\ \hline
\multirow{4}{*}{`high' (2)} & WG1         & 1.83      & 1.83           & 2.10             & 2.03               & 2.30                            & 2.30  \\
 & WG2\&3                                & 1.59        & 1.55           & 1.61          & 1.81               & 2.21                          & 2.39  \\
 & All reports                            & 1.66         & 1.64            & 1.76         & 1.88               & 2.24                       & 2.36 \\ 
 & Support                               & 98        & 99           & 100               & 100               & 100                          & 99 \\ \hline
\multirow{4}{*}{`medium' (1)} & WG1        & 1.45      & 1.35           & 1.71            & 1.49              & 2.26                            & 2.29  \\
 & WG2\&3                                & 1.19        & 1.20          & 1.31          & 1.37               & 2.18                          & 1.85  \\
 & All reports                            & 1.28         & 1.25            & 1.45          & 1.41               & 2.21                        & 2.00 \\ 
 & Support                                 & 99         & 99           & 100               & 100               & 100                           & 100 \\ \hline
\multirow{4}{*}{`low' (0)} & WG1          & 1.20       & 1.21           & 1.20             & 1.10               & 2.15                            & 1.70  \\
 & WG2\&3                                & 1.10        & 1.07           & 1.13          & 1.03               & 2.03                          & 1.90  \\
 & All reports                            & 1.14         & 1.12            & 1.16          & 1.06               & 2.08                        & 1.82 \\ 
 & Support                                  & 50        & 49           & 50                & 50               & 50                           & 49 \\ \hline
\multirow{3}{*}{Aggregate} & Ground truth & 1.50      & 1.50           & 1.50             & 1.50               & 1.50                            & 1.5  \\
 & Predicted                               & 1.46        & 1.44           & 1.56          & 1.60              & 2.21                          & 2.18  \\
 & Support (pred)                              & 295       & 295           & 300                        & 300               & 300                          & 297 \\ \hline
\end{tabular}%
}
\caption{Detailed results: Model average predicted confidence scores for each class within each setting.}
\label{tab:confidence-results}
\end{table}

\clearpage
\onecolumn
\section{Appendix: Classifier Results Table}\label{sec:classifier-results}

Table \ref{tab:classifier-results} presents the precision, recall, and F1 score for each classifier, as well as support (number of sentences for which the model answered with a valid confidence label). Note that the `very high' class and the `low' class each have 50 total sentences, while the `high' and `medium' classes each have 100, for a total of 300 sentences in the test set. 

\begin{table*}[!ht]
\centering
\resizebox{1\textwidth}{!}{%
\begin{tabular}{llcccccc}

\multicolumn{2}{l}{\textbf{Models}} & \multicolumn{2}{c}{\textbf{GPT-3.5-turbo}} & \multicolumn{2}{c}{\textbf{GPT-4}} & \multicolumn{2}{c}{\textbf{Cohere Command XL}}\\

\multicolumn{2}{l}{Setting} & Zero-shot & Few-shot & Zero-shot & Few-shot & Zero-shot & Few-shot \\ \hline \hline

\multirow{4}{*}{`very high'} & Precision & 0.500      & 0.476           & 0.428             & 0.375               & 0.221                            & 0.238  \\
 & Recall                                & 0.146        & 0.208           & 0.120          & 0.180               & 0.300                          & 0.592  \\
 & F1                                   & 0.226         & 0.290            & 0.188          & 0.243               & 0.254                        & 0.339 \\ 
 & Support                              & 48        & 48           & 50                & 50               & 50                          & 49 \\ \hline
\multirow{4}{*}{`high'} & Precision        & 0.504        &  0.485          & 0.472           & 0.475               & 0.332                         & 0.383   \\
 & Recall                               & 0.582         & 0.505            & 0.680           & 0.660             & 0.760                            & 0.546  \\
 & F1                                    & 0.540        & 0.495           & 0.557            & 0.552            & 0.462                          & 0.450  \\
 & Support                              & 98        & 99           & 100               & 100               & 100                          & 99 \\ \hline
\multirow{4}{*}{`medium'} & Precision      & 0.389        & 0.389            & 0.410          & 0.466               & 0.500                         & 0.0   \\
 & Recall                               & 0.636         & 0.616            & 0.570           & 0.610           & 0.010                          & 0.0  \\
 & F1                                    & 0.483        & 0.477           & 0.477             & 0.528          & 0.020                         & 0.0  \\ 
 & Support                                 & 99         & 99           & 100               & 100               & 100                           & 100 \\ \hline
        \multirow{4}{*}{`low'} & Precision   & 0.167      & 0.143           & 0.667             & 0.833               & 1.000                           & 0.353  \\
 & Recall                                    & 0.020        & 0.041           & 0.040         & 0.100              & 0.020                          & 0.245  \\
& F1                                        & 0.036        & 0.064           & 0.076          & 0.179            & 0.039                            & 0.289  \\ 
 & Support                                  & 50        & 49           & 50                & 50               & 50                           & 49 \\ \hline
\multirow{4}{*}{Aggregate} & \textbf{Accuracy}       & \textbf{0.434}      & \textbf{0.417}            & \textbf{0.443}          & \textbf{0.470}               & \textbf{0.310}                         & \textbf{0.320}   \\
 & Macro F1                            &  0.321      &0.331            & 0.324                 & 0.376                & 0.194                             & 0.270  \\
                 & Weighted F1          & 0.384         & 0.384           & 0.389                 & 0.430                & 0.209                           & 0.254  \\
 & Support                              & 295       & 295           & 300                        & 300               & 300                          & 297 \\ \hline
\end{tabular}%
}
\caption{Detailed results: Model classification performance results for the 3 models we assessed in both the zero shot and few shot setting. Reported metrics: accuracy, weighted and macro F1 score, and class-wise recall, precision, and F1 metrics.}
\label{tab:classifier-results}
\end{table*}

\clearpage
\onecolumn
\section{Appendix: Robustness Check}\label{sec:robustness_check}
\subsection{Baseline dataset}
As a robustness check for our approach, we evaluate LLM performance when classifying the confidence of statements outside of climate science and policy using the same experimental setup and analysis methods. 

We constructed a small baseline data set consisting of 337 sentences, which includes:

\begin{itemize}
    \item 100 nonsensical but grammatically correct and complete sentences from the ReCoGS dataset \cite{wu-etal-2023-recogs}
    \item 100 confirmed true factoid statements and 100 confirmed false statements from the FEVER dataset \cite{Thorne18Fever}
    \item 37 statements sourced from the Onion News dataset, which may be true, false, or fictional \cite{onion_dataset, onion}
\end{itemize}

\subsection{Prompting and methods}
In addition to using the \textsc{ClimateX} dataset, we also prompt the models to classify the confidence levels of statements from our baseline dataset. The purpose of this check is two-fold; (1) check that our experimental method can produce signals differentiating performance between different models and different source materials, and (2) benchmark how models perform on classifying statements unrestricted to the climate science topic from a variety of sources, as a basis for comparing trends that we find when evaluating LLMs on the \textsc{ClimateX} dataset. 

We use as similar a prompt as possible to the one used for querying the model with \textsc{ClimateX} statements, but remove references to climate science and the IPCC reports from the prompt. This minimizes the change in task setup while opening the model and task to a wider range of potential input statements. The complete prompt template is shown in Figure \ref{figure:baselinetemplate}. 

\subsection{Results and Discussion}
Although we do not have ground truth confidence labels for our baseline data set, we can use our implicit understanding of the source of each statement to determine appropriate model responses. Nonsensical sentences from the ReCOGS dataset are unverifiable and should elicit a no confidence label or ``low'' confidence response. Verifiably false sentences from the FEVER dataset should elicit a ``low'' confidence response (score: 0), while verifiably true sentences should elicit a ``high'' or ``very high'' confidence response (score: 2 or 3). Satirical or fictional sentences from the Onion dataset should elicit a ``low'' or no confidence label response, although some true statements may elicit a ``high'' or ``very high'' response.

The results show that: 

\begin{itemize}
\setlength\itemsep{0.1em}
\item Cohere's \texttt{Command-XL} is consistently very over-confident on statements from all sources, including verifiably false statements, nonsensical/unverifiable sentences, and likely fictional/satirical statements. 
\item \texttt{GPT-3.5-turbo} and \texttt{GPT-4} have relatively high rates (57\%, 25\%) of refusing to provide a confidence label when prompted with nonsensical ReCOGS sentences; in contrast, \texttt{Command-XL} always responds with a confidence label.
\item \texttt{GPT-3.5-turbo} and \texttt{GPT-4} are reasonably able to evaluate the truth of factoid statements from the FEVER dataset, shown by the low mean scores (0.7, 0.3) assigned to false statements and higher mean scores (1.8 and 1.9) assigned to true statements. \texttt{GPT-4} is slightly better, assigning lower mean scores to verifiably false statements.
\end{itemize}

\begin{figure}[!ht]
    \centering
    \includegraphics[width=0.7\linewidth]{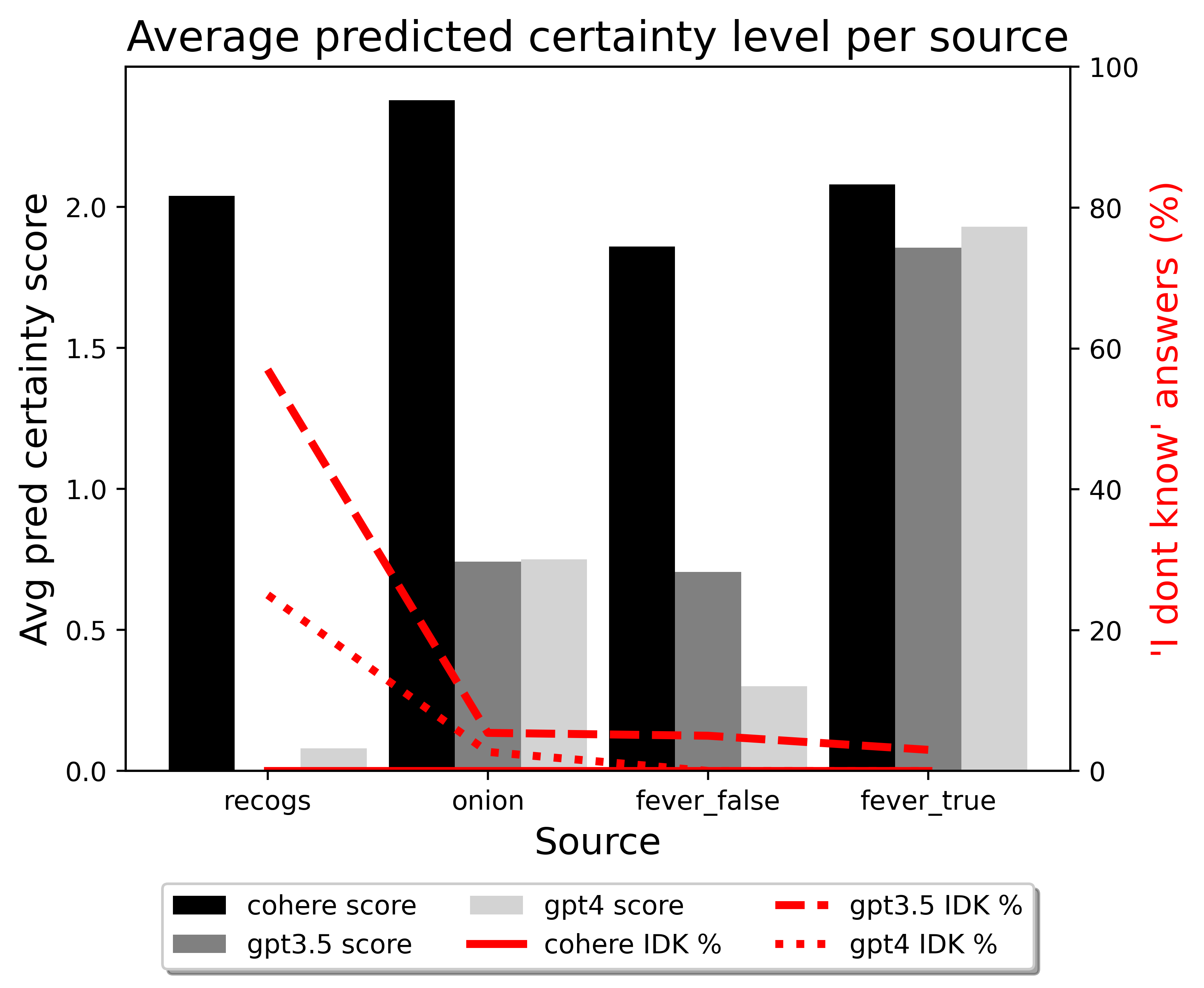}
    \caption{Mean predicted confidence score (left axis) and percentage of no confidence label responses (right axis) from each model, for each source within the robustness check dataset.}
    \label{fig:combined_baseline_results}
\end{figure}

Considering our climate-specific results in the context of baseline robustness check dataset, we see further evidence that some models can discern human confidence levels in specific statements. GPT-3 and \texttt{GPT-4} respond to requests for classifying nonsensical sentences from the ReCOGS dataset with no confidence label at a high rate, and can generally differentiate between true statements, false statements, and satirical/fictional statements appropriately. \textbf{Consistency in model performance on both our climate and baseline datasets suggest that our results are likely significant and not random}---it is likely that the models' classifications are based on true model ``knowledge'' from facts and confidence levels seen during training.

\subsection{Non-expert human robustness check}

Finally, as a baseline study, we performed a cursory non-expert human evaluation. We presented three college-educated (but non-scientist) volunteers with the sentences in the test set and tasked them with classifying the statements according to the 4 confidence classes, and obtained a 36.2\% accuracy. Further work is required to evaluate the performance of human experts on this task, and to create a more robust baseline among more volunteers.

\begin{figure}[!hb]
\capbtabbox{%
\resizebox{0.8\textwidth}{!}{%
\begin{tabular}{lcccccc}

 & Low  & Medium  & High  & Very high & Accuracy \\
 \hline \\
 
Non-expert humans & 1.46 & 1.50 & 1.89 & 1.88 & \textbf{36.2\%}
  \\ 
GPT-4 (few-shot) & 1.06 & 1.41 & 1.88 & 1.98 & \textbf{47.0\%}
 \\

\end{tabular}%
}}{%
\caption{Non-expert humans vs GPT-4 average confidence level predictions and overall accuracy.}
\label{tab:non-expert}
}
\end{figure}

\begin{figure}[htp]
\includegraphics[width=0.7\linewidth]{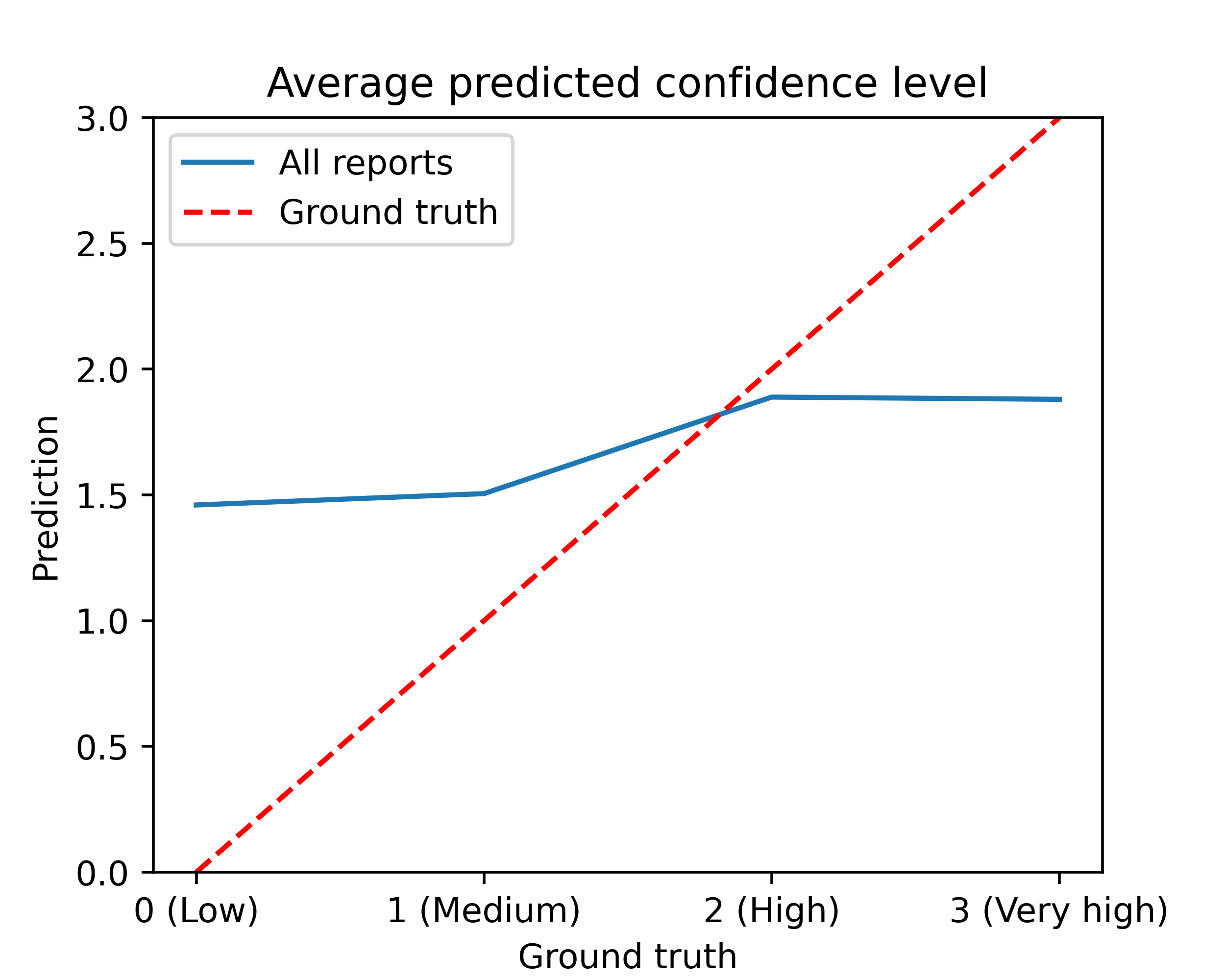}  

\caption{ Average confidence level predictions per ground truth class label for the non-expert human evaluation.}
\label{fig:non-expert}
\end{figure}

\clearpage
\onecolumn

\section{Appendix: Prompts}\label{sec:prompts}

\begin{figure*}[!ht]
\begin{mdframed}
You are a knowledgeable climate science assistant trained to assess the confidence level associated with various statements about climate change. \\
\\
You will be presented with a statement about climate science, climate impacts or climate change mitigation which is retrieved or paraphrased from the IPCC AR6 WGI, WGII or WGIII assessment reports. Climate scientists have evaluated that statement as low confidence, medium confidence, high confidence, or very high confidence, based on evidence (type, amount, quantity, consistency) and agreement among their peers. What is their confidence level? \\
\\
Respond *only* with one of the following words: 'low', 'medium', 'high', 'very high'. If you don't know, you can respond 'I don't know'. \\

---\\
\\
Follow the following format.\\
\\
Statement: \$\{a short statement about climate.\}\\
Confidence: \$\{must be *only*: 'low', 'medium', 'high', 'very high'\}\\
\\
---\\
\\
Statement: Since 1750, increases in CO2 (47\%) and CH4 (156\%) concentrations far exceed – and increases in N2O (23\%) are similar to – the natural multi-millennial changes between glacial and interglacial periods over at least the past 800,000 years\\
Confidence: \\
\end{mdframed}
\caption{Zero-shot template we used to prompt models for confidence levels associated with climate science statements from the \textsc{ClimateX} dataset, along with an example sentence.}
\label{figure:zeroshotprompt}
\end{figure*}

\begin{figure*}[!ht]

\begin{mdframed}
You are a knowledgeable assistant trained to assess the confidence level associated with various statements. \\     
\\
You will be presented with a statement. Humans have evaluated that statement as low confidence, medium confidence, high confidence, or very high confidence, based on evidence (type, amount, quantity, consistency) and agreement among their peers. What is their confidence level? \\
\\
Respond *only* with one of the following words: 'low', 'medium', 'high', 'very high'. If you don't know, you can respond 'I don't know'.\\
\\
---\\
\\
Follow the following format.\\
\\
Statement: \$\{a short statement.\} \\
Confidence: \$\{must be *only*: 'low', 'medium', 'high', 'very high'\} \\
\\
---\\
\end{mdframed}
\caption{Zero-shot template we used to prompt models for confidence levels associated with statements from our baseline dataset.}
\label{figure:baselinetemplate}
\end{figure*}

\end{document}